\documentclass{article}

\PassOptionsToPackage{numbers, compress}{natbib}

    \usepackage[final]{neurips_2023}

\usepackage{microtype}
\usepackage{graphicx}
\usepackage{booktabs} 
\usepackage{caption}
\usepackage{subcaption}

\usepackage[utf8]{inputenc} 
\usepackage[T1]{fontenc}    
\usepackage{hyperref}       
\usepackage{url}            
\usepackage{booktabs}       
\usepackage{amsfonts}       
\usepackage{nicefrac}       
\usepackage{microtype}      
\usepackage{xcolor}         

\usepackage{Jiazhi_style}
\usepackage{multirow}
\usepackage{enumitem}
\usepackage{wrapfig}
\usepackage{anyfontsize}

\usepackage{amsmath}
\usepackage{amssymb}
\usepackage{mathtools}
\usepackage{amsthm}

\title{Information-Theoretic Bounds on The Removal of Attribute-Specific Bias From Neural Networks}

\author{Jiazhi Li$^{1,2}$\thanks{Corresponding author: Jiazhi Li <jiazhil@usc.edu>} \quad Mahyar Khayatkhoei$^{1}$ \quad Jiageng Zhu$^{1,2}$ \quad Hanchen Xie$^{1,3}$ \\ \textbf{Mohamed E. Hussein}$^{1,4}$ \quad \textbf{Wael AbdAlmageed}$^{1,2,3}$ \\
$^1$USC Information Sciences Institute, Marina del Rey, USA \\
$^2$USC Ming Hsieh Department of Electrical and Computer Engineering, Los Angeles, USA \\
$^3$USC Thomas Lord Department of Computer Science, Los Angeles, USA \\
$^4$Alexandria University, Alexandria, Egypt
}

\begin{document}

\maketitle

\setcounter{footnote}{0}
\begin{abstract}
Ensuring a neural network is not relying on protected attributes (\eg race, sex, age) for predictions is crucial in advancing fair and trustworthy AI. While several promising methods for removing attribute bias in neural networks have been proposed, their limitations remain under-explored. In this work, we mathematically and empirically reveal an important limitation of attribute bias removal methods in presence of strong bias. Specifically, we derive a general non-vacuous information-theoretical upper bound on the performance of any attribute bias removal method in terms of the bias strength. We provide extensive experiments on synthetic, image, and census datasets to verify the theoretical bound and its consequences in practice. Our findings show that existing attribute bias removal methods are effective only when the inherent bias in the dataset is relatively weak, thus cautioning against the use of these methods in smaller datasets where strong attribute bias can occur, and advocating the need for methods that can overcome this limitation.
\end{abstract}

\section{Introduction}
\label{sec:introduction}
\emph{Protected attributes} is a term originating from Sociology~\cite{sociology} referring to a finite set of attributes that must not be used in decision-making to prevent exacerbating societal biases against specific demographic groups~\cite{protected_attributes}. For example, in deciding whether or not someone should be qualified for a bank loan, race (as one of the protected attributes)  must not influence the decision. Given the widespread use of neural networks in real-world decision-making, developing methods capable of explicitly excluding protected attributes from the decision process -- more generally referred to as removing attribute bias~\cite{minority_group_vs_sensitive_attribute} -- is of paramount importance.

While many promising methods for removing attribute bias in neural networks have been proposed in the recent years~\cite{BlindEye_IMDB_eb,learn_not_to_learn_Colored_MNIST,domain_independent_training, LfF_CelebA_Bias_conflicting,End,CSAD,BCL}, the limitations of these methods remain under-explored. In particular, existing studies explore the performance of these methods only in cases where the protected attribute (\eg race) is \emph{not strongly predictive} of the prediction target (\eg credit worthiness). However, this implicit assumption does not always hold in practice, especially in cases where training data is scarce. For example, in diagnosing Human Immunodeficiency Virus (HIV) from Magnetic Resonance Imaging (MRI), HIV subjects were found to be significantly older than control subjects, making age a strong attribute bias for this task~\cite{dataset_vs_task}. Another example is the Pima Indians Diabetes Database which contains only 768 samples where several spurious attributes become strongly associated with diabetes diagnosis~\cite{diabetes_dataset,diabetes_chapter}. Even the widely-used CelebA dataset~\cite{CelebA} contains strong attribute biases: for example in predicting blond hair, sex is a strong predictor~\footnote{We present detailed statistics of attribute biases in various real-world datasets in Appendix.}. Therefore, it is crucial to study the performance of bias removal methods beyond the moderate bias region to understand their limitations and the necessary conditions for their effectiveness.
 
\begin{figure}[t]
\begin{center}
  \includegraphics[width=0.9\linewidth]{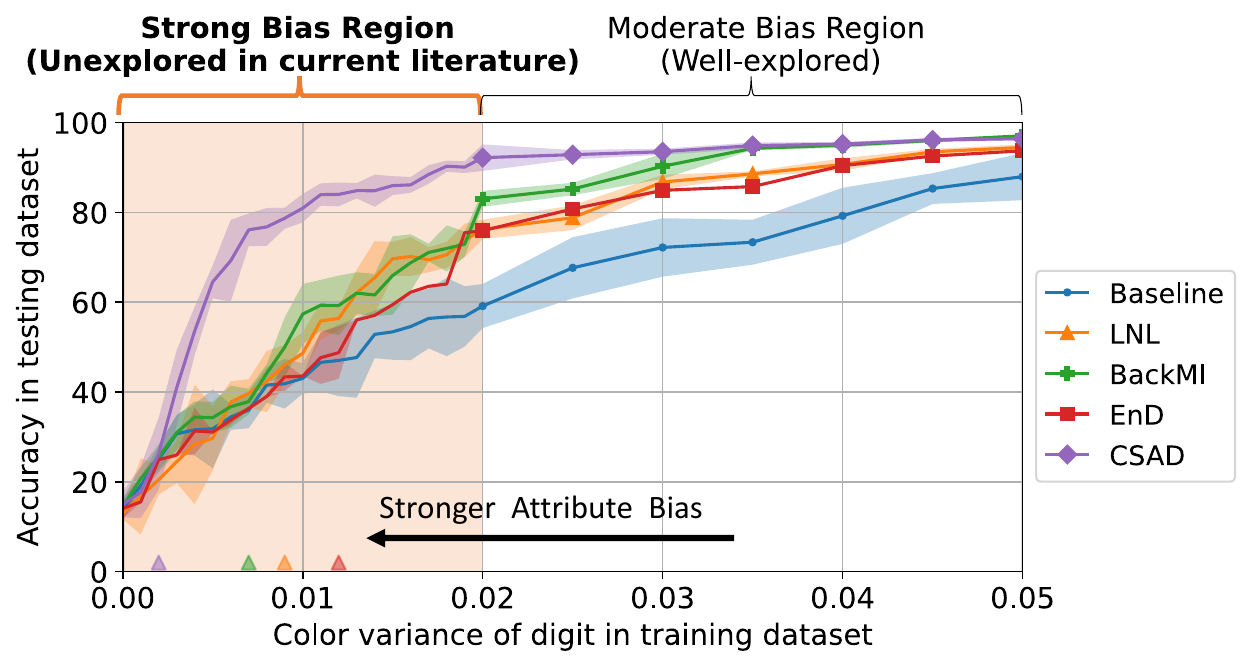}
\end{center}
  \caption{Accuracy of attribute bias removal methods under different levels of bias strength in Colored MNIST, showing results on the previously unexplored region of color variance $<0.02$. The breaking point of each method, where its performance becomes statistically similar to the baseline classifier, is labeled with $\blacktriangle$ on the x-axis. While all methods clearly outperform baseline in moderate bias region, their effectiveness sharply declines to baseline as bias strength increases. The plot shows average accuracy (lines) with one standard deviation error (shaded) over 15 randomized training runs.}
\label{fig:EB}
\end{figure}

In~\cref{fig:EB}, we will illustrate by a specific example the limitation in bias removal methods that we will later investigate theoretically and empirically in several real-world datasets. In this example, we conduct an extended version of a popular controlled experiment for evaluating the performance of attribute bias removal~\cite{learn_not_to_learn_Colored_MNIST, Back_MI, CSAD}. The task is to predict digits from colored MNIST images~\cite{learn_not_to_learn_Colored_MNIST} where color is considered a protected attribute. During training, each digit is assigned a unique RGB color with a variance (\ie the smaller the color variance, the more predictive the color is of the digit, and the stronger the attribute bias). To measure how much the trained model relies on the protected attribute in its predictions, model accuracy is reported on a held-out subset of MNIST with uniformly random color assignments (\ie where the color is not predictive of the digit).  While state-of-the-art  methods~\cite{learn_not_to_learn_Colored_MNIST,Back_MI,End,CSAD} report results for the color variance only in the range $[0.02, 0.05]$ (without providing any justification for this particular range), we explore the results for the missing range of $[0, 0.02]$, which we denote as the \emph{strong bias region}. As shown in~\cref{fig:EB}, in the strong bias region, we observe that the effectiveness of all existing methods sharply declines and that there exists a \emph{breaking point} in their effectiveness. The breaking point of a method is defined as the weakest bias strength at which its performance becomes indistinguishable from the baseline under a two-sample one-way Kolmogorov-Smirnov test with significance level of $0.05$. The main goal of this paper is to study the cause and extent of this limitation empirically and theoretically.

\section{Related Work}
\label{sec:related_work}
\textbf{Bias in Neural Networks.} Mitigating bias and improving fairness in neural networks has received considerable attention in recent years~\cite{EO_define,demographic_parity,counterfactual_fairness,fairness_through_awareness,fairness_under_unawareness,RLB}. The methods proposed for mitigating bias in neural networks can be broadly grouped into two categories: 1) methods that aim to mitigate the uneven performance of neural networks between majority and minority groups; and 2) methods that aim to reduce the dependence of neural network prediction on specific attributes. Most notable examples of the former group are methods for constructing balanced training set~\cite{Timnit_sex_PPB,Fairface}, synthesizing additional samples from the minority group~\cite{transect,CAT}, importance weighting the under-represented samples~\cite{RL_RBN}, and domain adaption techniques that adapt well-learnt representations from the majority group to the minority group~\cite{RFW,MFR,BAE}. In this work, our focus is on the second group of methods, which we will further divide into two subgroups discussed below: methods that implicitly or explicitly minimize the mutual information between learnt latent features and the specific protected attribute.

\noindent
\textbf{Explicit Mutual Information Minimization.} 
Several methods aim to directly minimize mutual information (MI) between a latent representation for the target classification and the protected attributes, in order to learn a representation that is predictive of the target but independent of the protected attributes, hence removing attribute bias. These methods mainly differ in the way they estimate MI. Most notable examples include LNL~\cite{learn_not_to_learn_Colored_MNIST} which minimizes the classification loss together with a MI regularization loss estimated by an auxiliary distribution; BackMI~\cite{Back_MI} which minimizes classification loss and MI estimated by a neural estimator~\cite{MINE} through the statistics network; and, CSAD~\cite{CSAD} which minimizes MI estimated by~\cite{deepinfomax} between a latent representation to predict target and another latent representation to predict the protected attributes.

\noindent
\textbf{Implicit Mutual Information Minimization.} 
Another group of methods aims to remove attribute bias by constructing surrogate losses that implicitly reduce the mutual information between protected attributes and the target of classification. Most notably, LfF~\cite{LfF_CelebA_Bias_conflicting} proposes training two models simultaneously, where the first model will prioritize easy features for classification by amplifying the gradient of cross-entropy loss with the predictive confidence (softmax score), and the second model will down-weight the importance of samples that are confidently classified by the first model, therefore avoiding features that are learnt easily during training, which are likely to be spurious features leading to large MI with protected attributes; EnD~\cite{End} adds regularization terms to the typical cross-entropy loss that push apart the feature vectors of samples with the same protected attribute label to become orthogonal (thereby increasing the conditional entropy of them given the protected attribute); BlindEye~\cite{BlindEye_IMDB_eb} pushes the distribution obtained by the attribute classifier operating on latent features towards uniform distribution by minimizing the entropy between them; and, domain independent training (DI)~\cite{domain_independent_training} learns a shared representation with an ensemble of separate classifiers per domain to ensure that the prediction from the unified model is not biased towards any domain.

\noindent
\textbf{Trade-offs between Bias Removal and Model Utility.}
The trade-offs between fairness and accuracy in machine learning models have garnered significant discussion.
Most notably, Kleinberg \etal~\cite{three_fairness_conditions} prove that except in highly constrained cases, no method can simultaneously satisfy three fairness conditions: \emph{calibration within groups} which requires that the expected number of individuals predicted as positive should be proportional to a group-specific fraction of individuals in each group, \emph{balance for the negative class} which requires that the average score of individuals predicted as negative should be equal across groups, and \emph{balance for the positive class} which requires the balance for the positive class across groups; and, Dutta \etal~\cite{Eopps_and_accuracy} theoretically demonstrate that, under certain conditions, it is possible to simultaneously achieve optimal accuracy and fairness in terms of \emph{equal opportunity}~\cite{EO_define} which requires even false negative rates or even true positive rates across groups.
Different from the above-mentioned fairness criteria, we focus on another well-known fairness criterion, \emph{demographic parity}~\cite{counterfactual_fairness, fairness_through_awareness}, which requires even prediction probability across groups, \ie independence between model prediction and protected attributes.
Regarding this criterion, Zhao and Gordon~\cite{dp_to_ap} show that any method designed to learn fair representations, while ensuring model predictions are independent of protected attributes, faces an information-theoretic lower bound on the joint error across groups.
In contrast, we derive a general information-theoretic upper bound on the best attainable performance, which is not limited to the case where model predictions are independent of protected attributes and considers different levels of the retained protected attribute information in the learnt features.

\section{Bounding the Performance of Attribute Bias Removal Methods}
\begin{figure*}[h]
    \centering 

      \subfloat[Baseline.]{\includegraphics[width=0.3\linewidth]{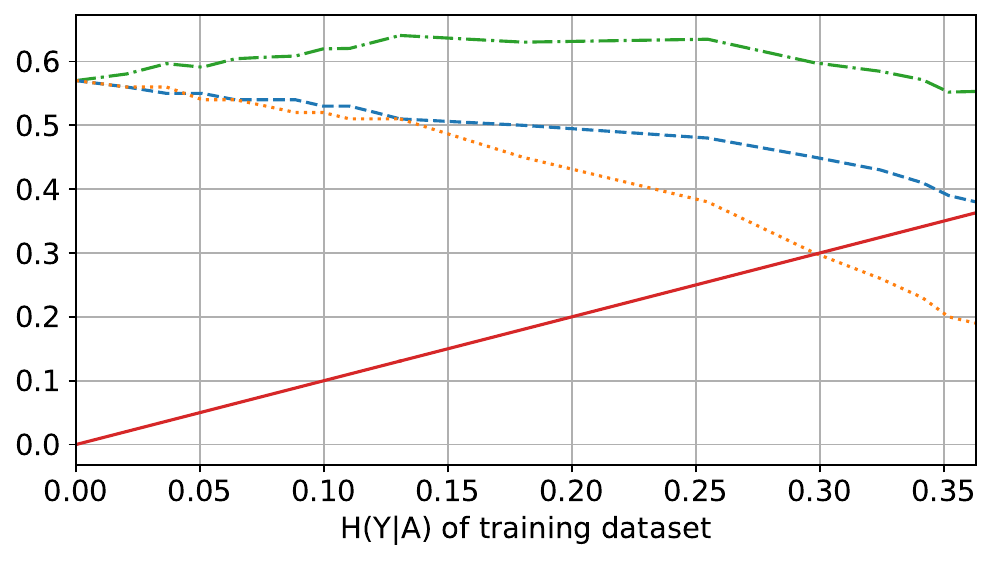}\label{fig:CelebA_baseline_bound_supp}}\quad
      \subfloat[LNL~\cite{learn_not_to_learn_Colored_MNIST}.]{\includegraphics[width=0.3\linewidth]{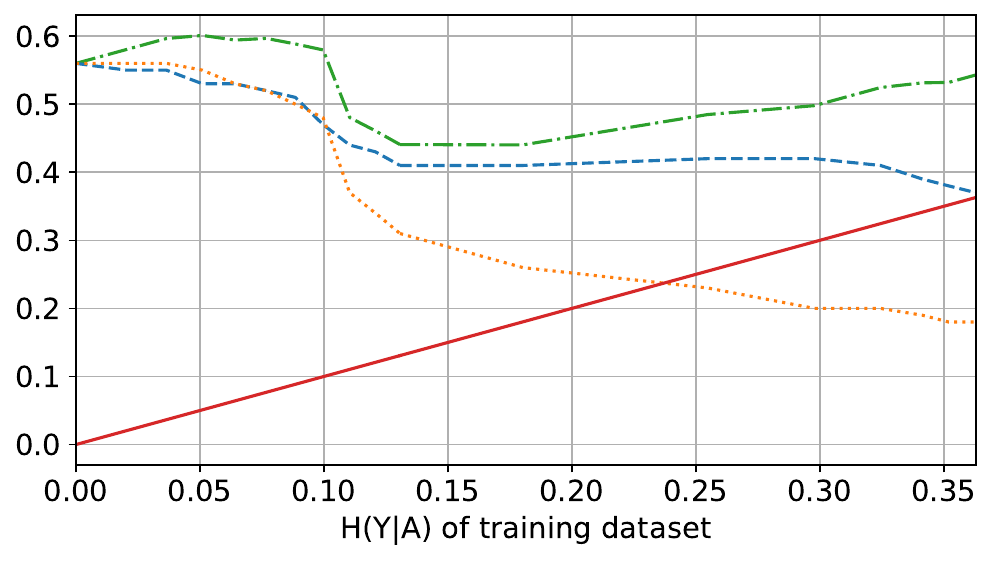}\label{fig:CelebA_LNL_bound_supp}}\quad
      \subfloat[DI~\cite{domain_independent_training}.]{\includegraphics[width=0.3\linewidth]{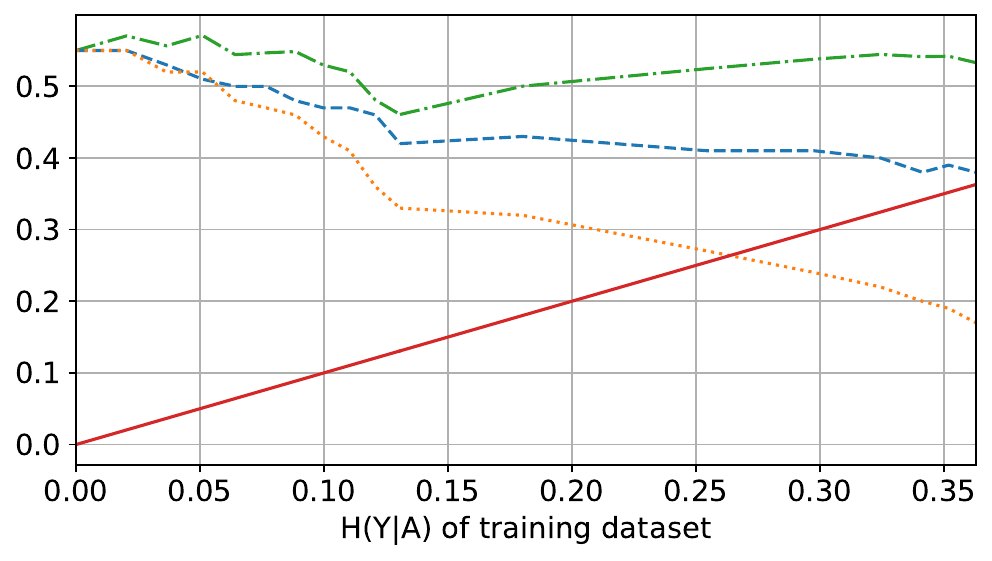}\label{fig:CelebA_DI_bound_supp}}

      \subfloat[LfF~\cite{LfF_CelebA_Bias_conflicting}.]{\includegraphics[width=0.3\linewidth]{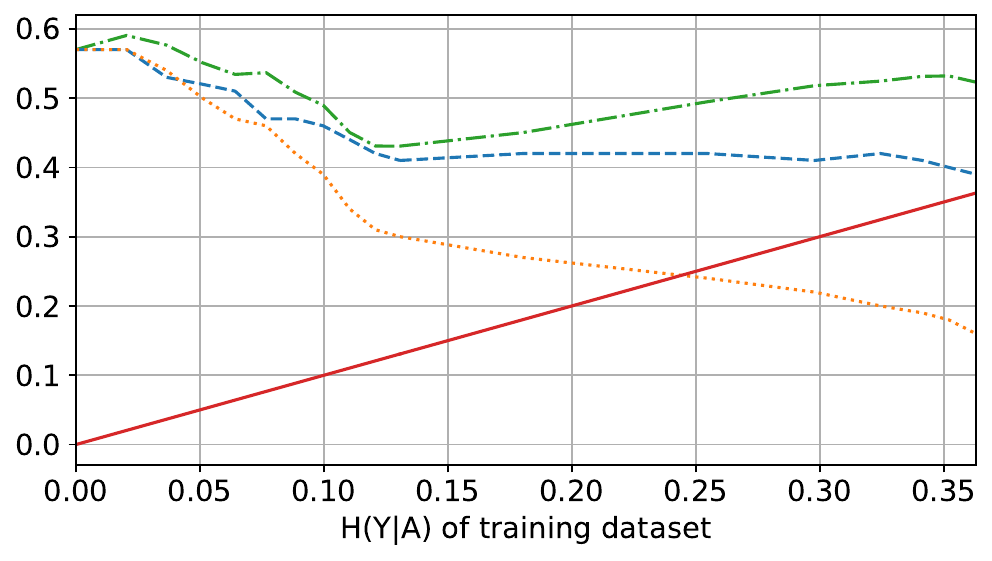}\label{fig:CelebA_LfF_bound_supp}}\quad
      \subfloat[EnD~\cite{End}.]{\includegraphics[width=0.3\linewidth]{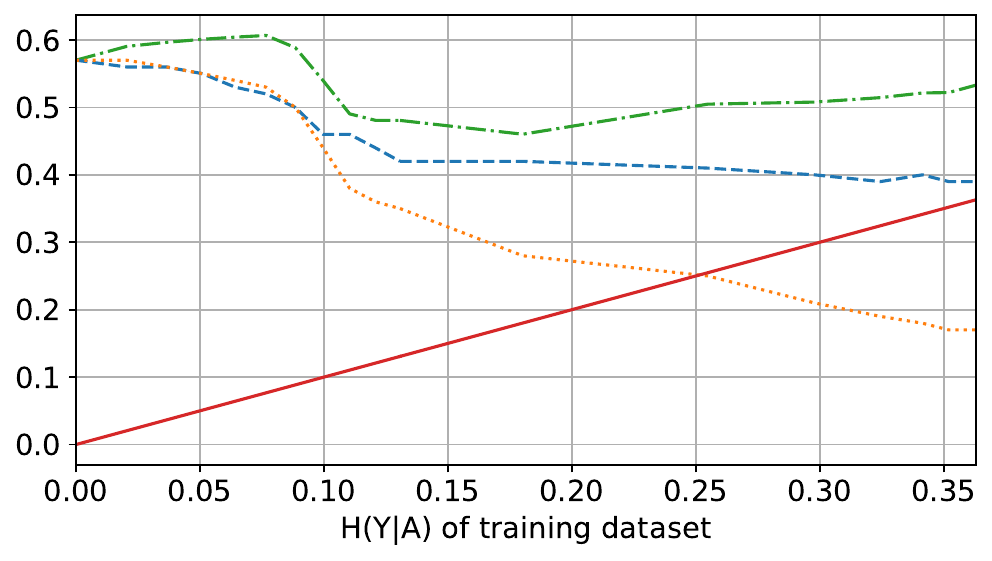}\label{fig:CelebA_EnD_bound_supp}}\quad
      \subfloat[CSAD~\cite{CSAD}.]{\includegraphics[width=0.3\linewidth]{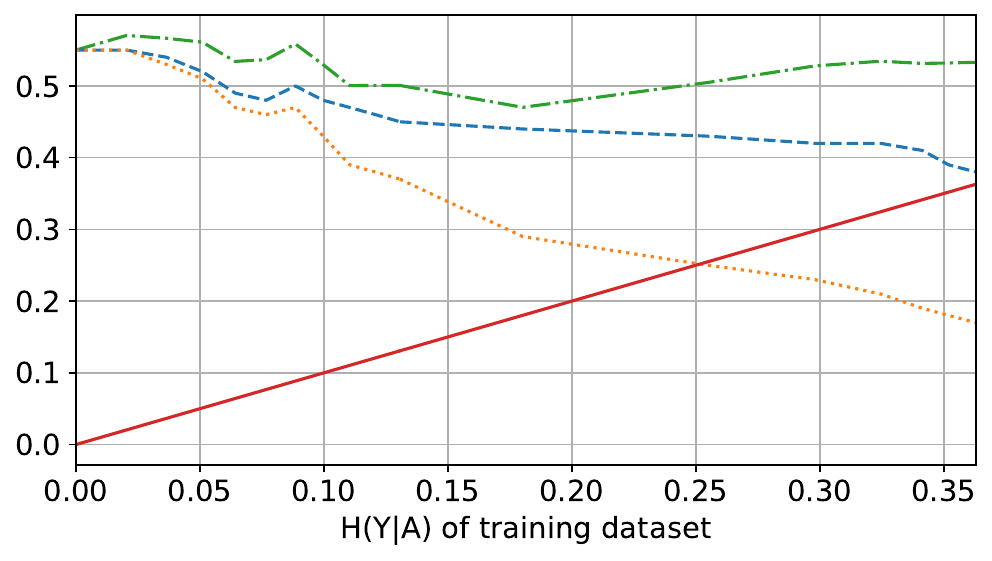}\label{fig:CelebA_CSAD_bound_supp}}

      \subfloat[BCL~\cite{BCL}.]{\includegraphics[width=0.9\linewidth]{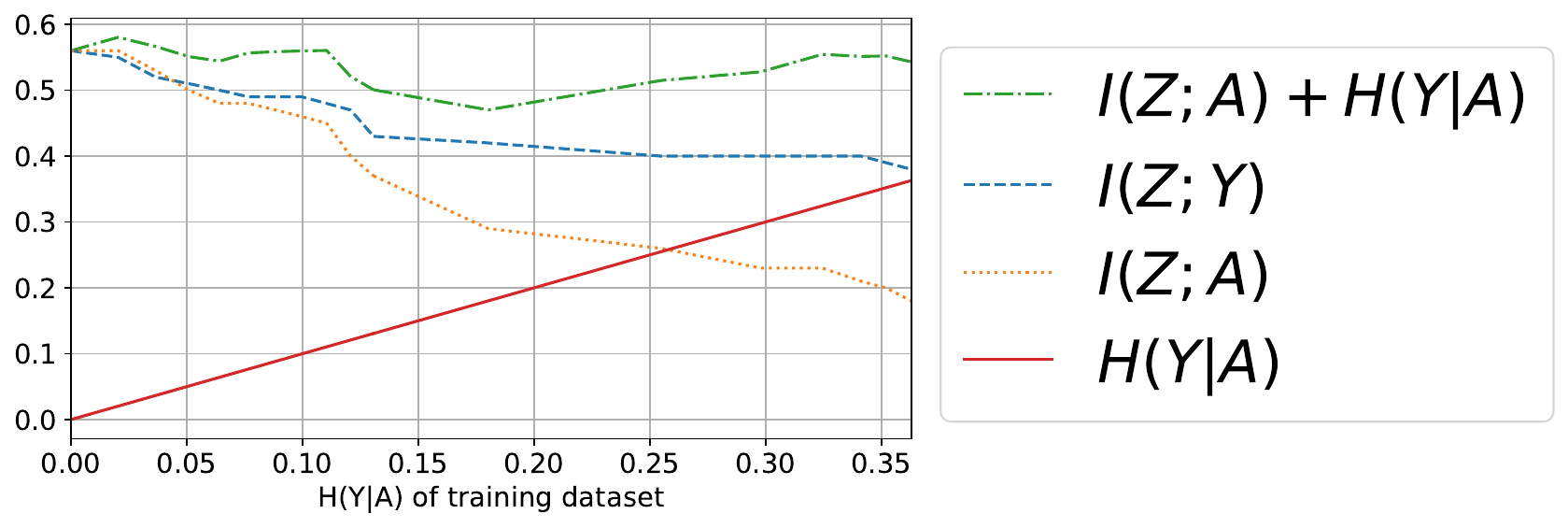}\label{fig:CelebA_BCL_bound_supp}}\quad

\caption{Empirically verifying the bound in~\cref{th:general} on CelebA. 
The x-axis shows $H(Y|A)$, which we vary directly by adjusting the fraction of bias-conflicting images while ensuring a constant number of biased images in the training set. We empirically compute $H(Y|A)$ based on the distribution of $Y$ and $A$ in the modified training set, and estimate mutual information using~\cite{MINE}. The bound $0 \leq I(Z;Y) \leq I(Z;A) + H(Y|A)$ holds in all cases.
}
\label{fig:CelebA_bounds_supp}
\end{figure*}
\label{sec:theory}
The observations in~\cref{fig:EB} revealed that the existing methods are not effective when the attribute bias is too strong, \ie they all have a breaking point, and that there is a continuous connection between their effectiveness and the strength of the attribute bias. However, so far, these observations are limited to the particular Colored MNIST dataset. In this section, we show that this situation is in fact much more general. By deriving an upper bound on the classification performance of any attribute bias removal method in terms of the bias strength, regardless of the dataset and domain, we will elucidate the cause and extent of the limitation we observed in~\cref{fig:EB}.

We first need to formalize the notions of performance, attribute bias strength, and attribute bias removal. Let $X$ be a random variable representing the input (\eg images) with support $\mathcal{X}$, $Y$ a random variable representing the prediction target (\eg hair color) with support $\mathcal{Y}$, and $A$ a random variable representing the protected attribute (\eg sex). We define the attribute bias removal method as a function $f: \mathcal{X} \rightarrow \mathcal{Z}$ that maps input data to a latent bottleneck feature space $\mathcal{Z}$ inducing the random variable $Z$, and consider the prediction model as a function $g: \mathcal{Z} \rightarrow \mathcal{Y}$ inducing the random variable $\hat{Y}$. According to the information bottleneck theory~\cite{tishby2015deep-bottleneck-theory, shwartz2017opening-blackbox-deep-learning}, the goal of classification can be stated as maximizing the mutual information between prediction and target, namely $I(\hat{Y};Y)$, which is itself bounded by the mutual information between feature and target due to the data processing inequality~\cite{cover2006elements}, \ie $I(\hat{Y};Y) \leq I(Z;Y)$. Intuitively, $I(Z;Y)$ measures how informative the features learnt by the model are of the target, with $I(Z;Y)=0$ indicating completely uninformative learnt features: the best attainable prediction performance is no better than chance. Therefore, the optimization objective of attribute bias removal methods can be formalized as learning $f$ parameterized by $\theta$ that minimizes mutual information between feature and attribute $I(Z_{\theta};A)$, while maximizing mutual information between feature and target $I(Z_{\theta};Y)$.

Given the above definitions, we can state our goal in this section concretely: to derive a connection between $I(Z;Y)$ (the best attainable performance), $H(Y|A)$ (the attribute bias strength measured by the conditional entropy of target given attribute), and $I(Z;A)$ (the amount of remaining attribute bias in the learnt feature). 
Note that the stronger the attribute bias is, the better the attribute can predict the target, hence the lower $H(Y|A)$. So the extreme attribute bias happens when $H(Y|A)=0$. In this particular extreme setting, the following proposition shows that no classifier can outperform random guess if the attribute is removed from the feature, \ie $I(Z;A)=0$.
\begin{propos}
\label{th:extreme}
Given random variables $Z, Y, A$, in case of the extreme attribute bias $H(Y|A)=0$, if the attribute is removed from the feature $I(Z;A)=0$, then $I(Z;Y)=0$, \ie no classifier can outperform random guess.~\footnote{\label{fnote_proof}Proof in Appendix.}
\end{propos}

This proposition extends and explains the observation on the leftmost location of the x-axis in~\cref{fig:EB}: when the color variance is zero, color is completely predictive of the digit, $H(Y|A)=0$, and removing color from the latent feature, $I(Z;A)=0$, makes the prediction uninformative, $I(Z;Y)=0$. However, \cref{th:extreme} does not explain the rest of the curve beyond just zero color variance. The following theorem closes this gap by deriving a bound on the performance of attribute bias removal methods in terms of the attribute bias strength, thus providing a more complete picture of the limitation of attribute bias removal, and elucidating the connection of performance and bias strength.
\begin{theorem}
\label{th:general}
Given random variables $Z, Y, A$, the following inequality holds without exception:~\cref{fnote_proof}
\begin{align}
\label{eq:genreal}
    0 \leq I(Z;Y) \leq I(Z;A) + H(Y|A)
\end{align}
\end{theorem}

\begin{remark}
In the extreme bias case $H(Y|A)=0$, the bound in~\cref{eq:genreal} shows that the model performance is bounded by the amount of protected attribute information that is retained in the feature, namely $I(Z;Y) \leq I(Z;A)$. This puts the model in a trade-off: the more the attribute bias is removed, the lower the best attainable performance.
\end{remark}

\begin{remark}
When the protected attribute is successfully removed from the feature $I(Z;A)=0$, the bound in~\cref{eq:genreal} shows that the model's performance is bounded by the strength of the attribute bias, namely $I(Z;Y)\leq H(Y|A)$. This explains the gradual decline observed in~\cref{fig:EB} as we moved from the moderate to the strong bias region (from right to left towards zero color variance).
\end{remark}

\begin{remark}
When $H(Y|A)=0$ and $I(Z;A)=0$, \cref{eq:genreal} reduces to the result of~\cref{th:extreme}, $I(Z;Y)=0$, hence no classifier can outperform random guess.
\end{remark}

\begin{remark}
\label{th:best_att_perf}
We emphasize that the provided bound is placed on the best attainable performance. So while decreasing the bound will decrease performance, increasing the bound will not necessarily result in an increased performance. For example, consider the baseline classifier: even though there is no attribute bias removal performed and therefore the bound can be arbitrarily large, $I(Z;A) \gg 0$, the model declines in the strong bias region since learning the highly predictive protected attribute is likely in the non-convex optimization.
\end{remark}

To empirically test our theory in a real-world dataset, we compute the terms in~\cref{th:general} for several attribute bias removal methods in CelebA and plot the results in~\cref{fig:CelebA_bounds_supp}. In these experiments, blond hair is the target $Y$, and sex is the protected attribute $A$. We vary the bias strength $H(Y|A)$ by increasing/decreasing the fraction of bias-conflicting images in the training set (images of females with non-blond hair and males with blond hair) while maintaining the number of biased images in the training set at 89754. Then, we compute $H(Y|A)$ directly and estimate the mutual information terms $I(Z;A)$ and $I(Z;Y)$ using mutual information neural estimator~\cite{MINE}. We observe that the bound holds in accordance with~\cref{th:general} in all cases. Now that we have mathematically and empirically shown the existence of the bound, in the next section, we will investigate the extent of its consequences for attribute bias removal methods in image and census datasets, in addition to the consequence we have already observed in the synthetic Colored MNIST dataset in~\cref{fig:EB}.

\section{Experiments}
\label{sec:exp}
In this section, we empirically study the performance of the existing attribute bias removal methods in the strong bias setting. We conduct experiments with an extensive list of existing state-of-the-art attribute bias removal methods~\cite{learn_not_to_learn_Colored_MNIST,domain_independent_training, LfF_CelebA_Bias_conflicting, End, CSAD, BCL} on Colored MNIST as well as two real-world datasets: CelebA~\cite{CelebA} and Adult~\cite{UCI_ML_Repo}. 
For all results, we report average performance with one standard deviation over multiple trials (15 trials in Colored MNIST, 5 in CelebA, 25 in Adult). 

\noindent
\textbf{Colored MNIST Dataset}
is an image dataset of handwritten digits, where each digit is assigned a unique RGB color with a certain variance, studied by these methods~\cite{learn_not_to_learn_Colored_MNIST,Back_MI,End,CSAD}. The training set consists of 50000 images and the testing set of 10000 images with uniformly random color assignment. The color is considered the protected attribute $A$ and the digit is the target $Y$. The variance of color in the training set determines the strength of the bias $H(Y|A)$. The results on this dataset are reported in~\cref{fig:EB} and explained in~\cref{sec:introduction}.

\noindent
\textbf{CelebA Dataset}~\cite{CelebA} is an image dataset of human faces studied by these methods~\cite{learn_not_to_learn_Colored_MNIST,domain_independent_training,LfF_CelebA_Bias_conflicting,End,CSAD,BCL}. Facial attributes are considered the prediction target $Y$ (\eg blond hair), and sex is the protected attribute $A$. For each target, there is a notion of \textit{biased samples} -- images in which $Y$ is positively correlated with $A$, \eg images of females with blond hair and males without blond hair -- and a notion of \textit{bias-conflicting} samples -- images in which $Y$ is negatively correlated with $A$, \eg images of females without blond hair and males with blond hair. The fraction of bias-conflicting images in the training set determines the strength of the bias $H(Y|A)$. For training, we consider the original training set of CelebA denoted \textit{TrainOri} consisted of 162770 images with $H(Y|A)=0.36$, and an extreme bias version in which the bias-conflicting samples are removed from the original training set denoted \textit{TrainEx} consisted of 89754 images with $H(Y|A)=0$. Additionally, we construct 16 training sets between TrainOri and TrainEx by maintaining the number of biased samples and varying the fraction of bias-conflicting samples. For testing, we consider two versions of the original testing set: (1) \emph{Unbiased} consists of 720 images in which all pairs of target and protected attribute labels have the same number of samples, and (2) \emph{Bias-conflicting} consists of 360 images in which biased samples are excluded from the \emph{Unbiased} dataset (only bias-conflicting samples remain). 

\noindent
\textbf{Adult Dataset}~\cite{UCI_ML_Repo} is a census dataset of income which is a well-known fairness benchmark. Income is considered the target $Y$ and sex is the protected attribute $A$. To construct training and testing sets, we follow the setup of CelebA explained above, but we further mitigate the effect of data imbalance and the variation in the total number of training samples. For training, we consider the balanced version of the original training set of Adult denoted \textit{TrainOri} consisted of 7076 records with $H(Y|A)=0.69$, and an extreme bias version in which the bias-conflicting samples are removed from TrainOri and the same number of biased samples are appended denoted \textit{TrainEx} with $H(Y|A)=0$ consisted of the same total number (7076) of records as TrainOri. Additionally, we construct 11 training sets in between TrainOri and TrainEx by varying the fraction of biased samples in TrainEx while maintaining the total size of training set. For testing, we consider two versions of the original testing set: (1) \emph{Unbiased} consists of 7076 records in which all pairs of target and protected attribute labels have the same number of samples, and (2) \emph{Bias-conflicting} consists of 3538 records in which biased samples are excluded from the \emph{Unbiased} dataset (only bias-conflicting samples remain).

\noindent
\textbf{Training Details.}
For digit classification on Colored MNIST, we use Lenet-5~\cite{lenet} as the baseline classifier.
For facial attribute classification on CelebA, following CSAD~\cite{CSAD}, we use ResNet-18~\cite{ResNet} as the baseline classifier. 
For income prediction on Adult dataset, we use a three-layer MLP as the baseline classifier. For all experiments, we set learning rate to 0.001, batch size to 32, and Adam optimizer with $\beta_1=0.9$ and $\beta_2=0.999$. All hyperparameters are set according to respective papers.

\begin{table*}[t]
\caption{Performance of attribute bias removal methods trained under extreme bias in CelebA (\emph{TrainEx} training set) to predict \textit{blond hair}. $\Delta$ indicates the difference from baseline. None of the methods can effectively remove the bias $I(Z;A)$ compared to baseline.}
\label{tab:CelebA_BlondHair}
\centering

\begin{tabular}{lcccc}
\toprule
\multirow{2}{*}{Method}          & \multicolumn{2}{c}{Test Accuracy}         & \multicolumn{2}{c}{Mutual Information} \\
\cmidrule(lr){2-3}  \cmidrule(lr){4-5} 
                                 & Unbiased ↑          & Bias-conflicting ↑  & $I(Z;A)$ ↓          & $\Delta$ (\%) ↑  \\
                                 \midrule
Random guess                          & 50.00  & 50.00 & 0.57 & 0.00             \\
Baseline                          & 66.11{\scriptsize $\pm$0.32} & 33.89{\scriptsize $\pm$0.45} & 0.57{\scriptsize $\pm$0.01} & 0.00             \\
\midrule
LNL~\cite{learn_not_to_learn_Colored_MNIST}                               & 64.81{\scriptsize $\pm$0.17} & 29.72{\scriptsize $\pm$0.26} & 0.56{\scriptsize $\pm$0.06} & 1.75             \\
DI~\cite{domain_independent_training}                                & 66.83{\scriptsize $\pm$0.44} & 33.94{\scriptsize $\pm$0.65} & 0.55{\scriptsize $\pm$0.02} & 3.51             \\
LfF~\cite{LfF_CelebA_Bias_conflicting}                               & 64.43{\scriptsize $\pm$0.43} & 30.45{\scriptsize $\pm$1.63} & 0.57{\scriptsize $\pm$0.03} & 0.00             \\
EnD~\cite{End}                               & 66.53{\scriptsize $\pm$0.23} & 31.34{\scriptsize $\pm$0.89} & 0.57{\scriptsize $\pm$0.05} & 0.00             \\
CSAD~\cite{CSAD}                              & 63.24{\scriptsize $\pm$2.36} & 29.13{\scriptsize $\pm$1.26} & 0.55{\scriptsize $\pm$0.04} & 3.51             \\
BCL~\cite{BCL}                               & 65.30{\scriptsize $\pm$0.51} & 33.44{\scriptsize $\pm$1.31} & 0.56{\scriptsize $\pm$0.07} & 1.75             \\
\bottomrule
\end{tabular}

\end{table*}

\begin{table*}[t]
        \caption{Performance of attribute bias removal methods trained under extreme bias in Adult (\emph{TrainEx} training set) to predict income. $\Delta$ indicates the difference from baseline. None of the methods can effectively remove the bias $I(Z;A)$ compared to baseline.}
        \label{tab:Adult_mostEx}
        \centering
\begin{tabular}{lcccc}
\toprule
\multirow{2}{*}{Method}  & \multicolumn{2}{c}{Test Accuracy}         & \multicolumn{2}{c}{Mutual Information} \\
\cmidrule(lr){2-3}  \cmidrule(lr){4-5} 
                         & Unbiased ↑          & Bias-conflicting ↑  & $I(Z;A)$ ↓       & $\Delta$ (\%) ↑      \\
                         \midrule
Random guess             & 50.00                  & 50.00                   & 0.69             & 0.00                \\
Baseline                 & 50.59{\scriptsize $\pm$0.54}          & 1.19{\scriptsize $\pm$0.83}           & 0.69{\scriptsize $\pm$0.00}        & 0.00                \\
\midrule
LNL~\cite{learn_not_to_learn_Colored_MNIST}                      & 50.10{\scriptsize $\pm$0.18}          & 0.43{\scriptsize $\pm$0.46}           & 0.69{\scriptsize $\pm$0.01}        & 0.00                \\
DI~\cite{domain_independent_training}                       & 50.61{\scriptsize $\pm$0.28}          & 0.65{\scriptsize $\pm$0.64}           & 0.69{\scriptsize $\pm$0.01}        & 0.00                \\
LfF~\cite{LfF_CelebA_Bias_conflicting}                      & 50.33{\scriptsize $\pm$0.34}          & 0.78{\scriptsize $\pm$0.65}           & 0.69{\scriptsize $\pm$0.01}        & 0.00                \\
EnD~\cite{End}                      & 50.59{\scriptsize $\pm$0.75}          & 1.18{\scriptsize $\pm$0.96}           & 0.69{\scriptsize $\pm$0.00}        & 0.00                \\
CSAD~\cite{CSAD}                     & 50.76{\scriptsize $\pm$2.22}          & 1.43{\scriptsize $\pm$2.46}           & 0.69{\scriptsize $\pm$0.01}        & 0.00                \\
BCL~\cite{BCL}                      & 50.83{\scriptsize $\pm$1.34}          & 0.52{\scriptsize $\pm$0.83}           & 0.69{\scriptsize $\pm$0.00}        & 0.00                \\
\bottomrule
\end{tabular}
\end{table*}

\begin{figure*}[b!]
  \centering

      \subfloat[Accuracy of \emph{Unbiased} testing set.]{\includegraphics[width=0.7\linewidth]{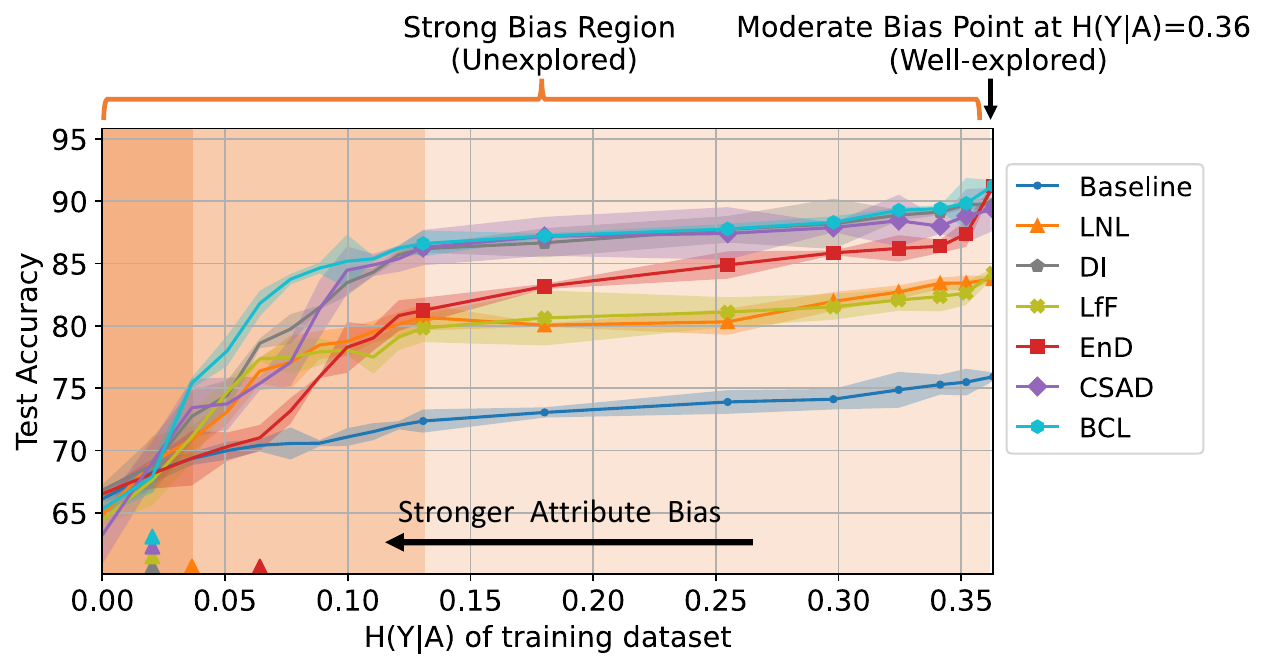}\label{fig:CelebA_Unbiased_Acc}}\quad
      \subfloat[Accuracy of \emph{Bias-conflicting} testing set.]{\includegraphics[width=0.45\linewidth]{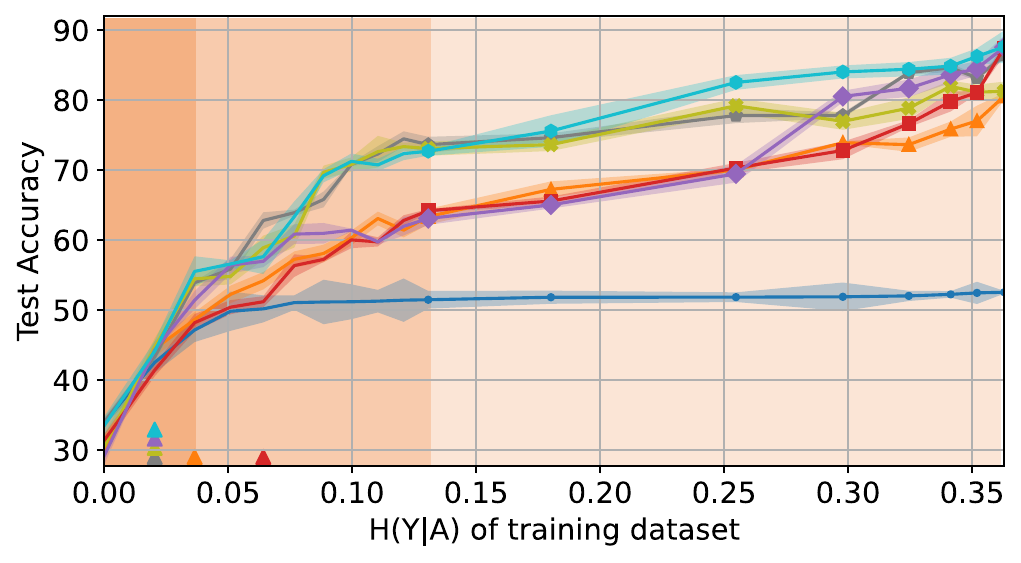}\label{fig:CelebA_conflict_Acc}}\quad
      \subfloat[Attribute bias $I(Z;A)$ of training set.]{\includegraphics[width=0.45\linewidth]{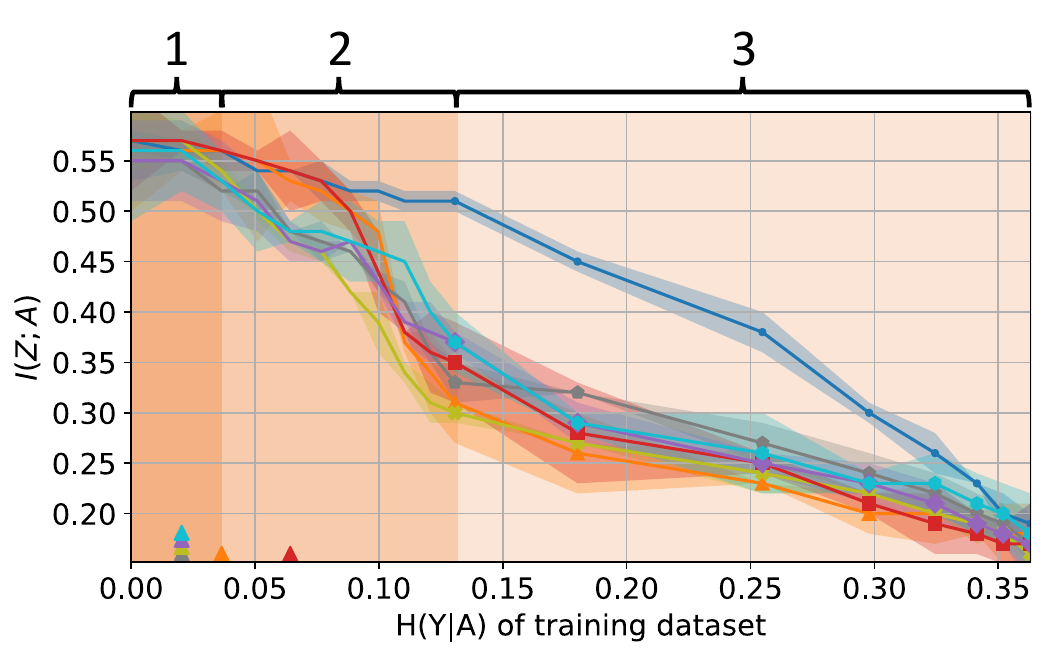}\label{fig:CelebA_IZA}}\quad
  
  \caption{Accuracy and mutual information under different bias strengths in CelebA. As bias strength increases (moving from right to left), the performance of all methods degrades and sharply declines to baseline at the breaking point (labeled by $\blacktriangle$).}
  \label{fig:EB_CelebA}
\end{figure*}
\begin{figure*}[ht]
  \centering

      \subfloat[Accuracy of \emph{Unbiased} testing set.]{\includegraphics[width=0.7\linewidth]{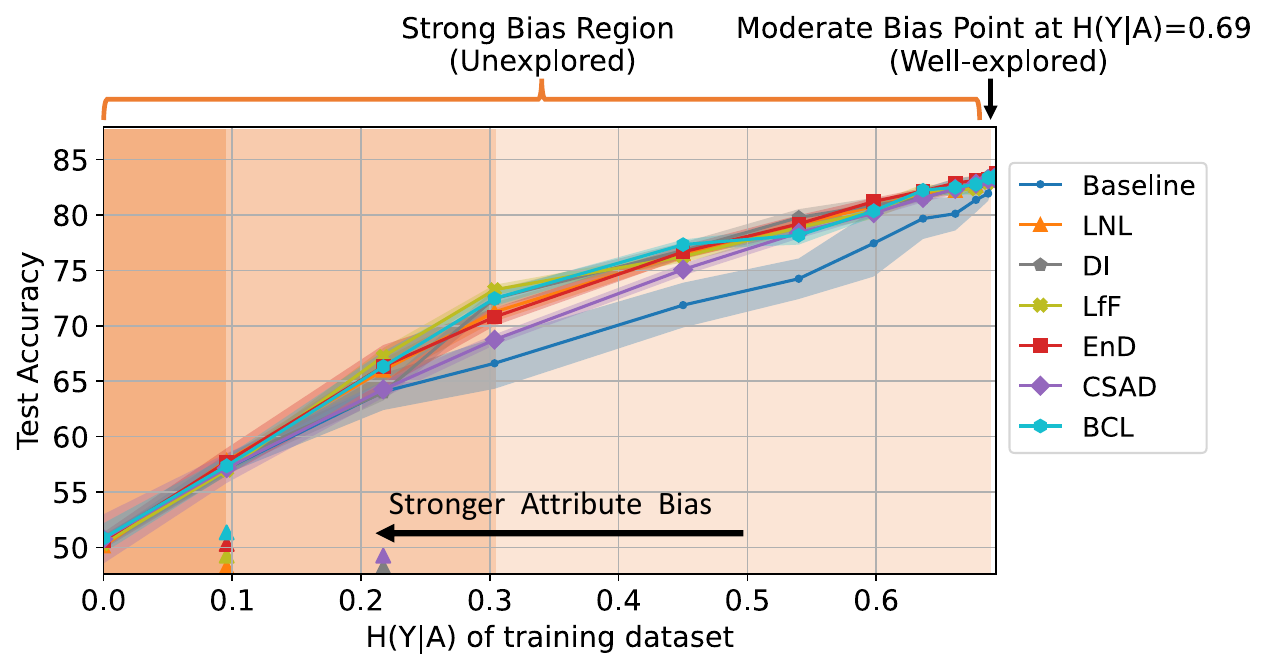}\label{fig:Adult_Unbiased_Acc}}\quad
      \subfloat[Accuracy of \emph{Bias-conflicting} testing set.]{\includegraphics[width=0.45\linewidth]{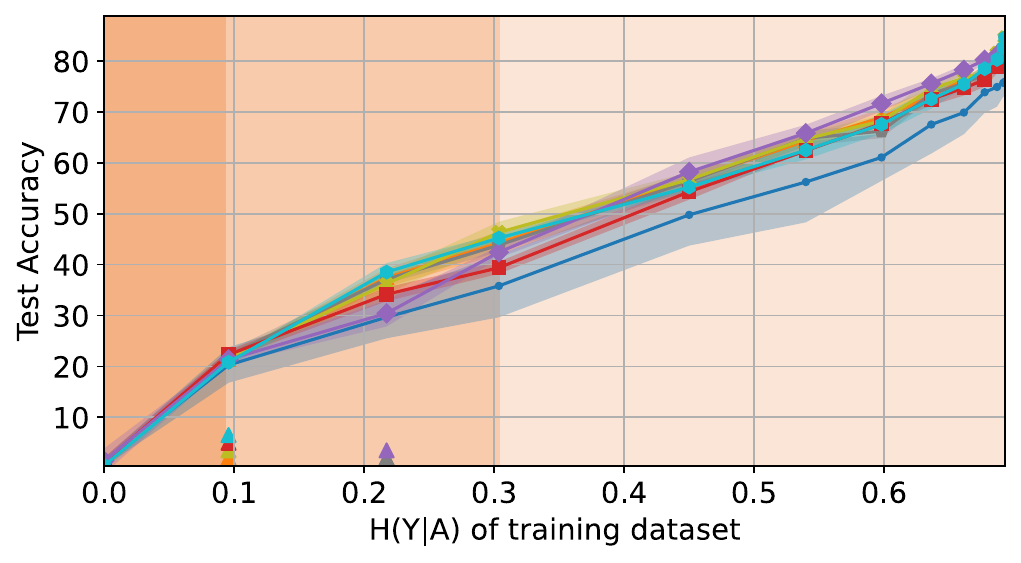}\label{fig:Adult_conflict_Acc}}\quad
      \subfloat[Attribute bias $I(Z;A)$ of training set.]{\includegraphics[width=0.45\linewidth]{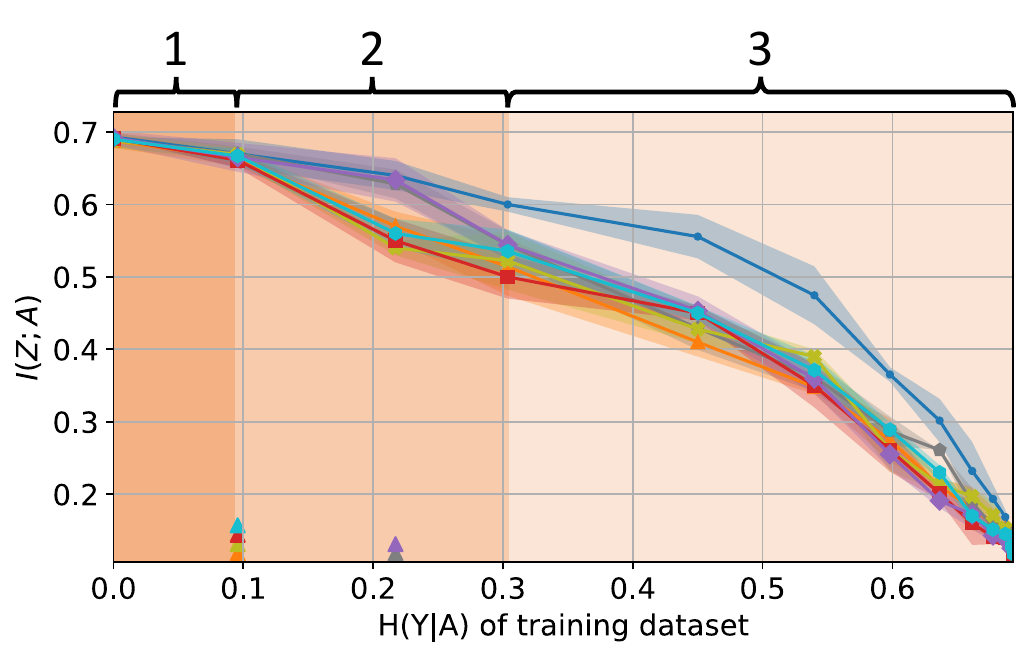}\label{fig:Adult_IZA}}\quad

  \caption{Accuracy and mutual information under different bias strengths in Adult. As bias strength increases (moving from right to left), the performance of all methods degrades and sharply declines to baseline at the breaking point (labeled by $\blacktriangle$).}
  \label{fig:EB_Adult}
\end{figure*}

\subsection{Analysis of the Extreme Bias Point $H(Y|A)=0$}
\label{sec:comp_extreme}
In this section, we investigate the consequences of applying existing attribute bias removal methods at the extreme bias point $H(Y|A)=0$. We study two aspects of each method, the classification performance (measured by accuracy on Unbiased and Bias-conflicting settings) and its ability to remove bias (measured by estimating $I(Z;A)$ using~\cite{MINE} on the training set). Ideally, a method must achieve on-par or better accuracy than the baseline while learning a representation $Z$ that does not reflect the attribute bias present in the training set ($I(Z;A)=0$), hence successfully removing the bias. However, in~\cref{tab:CelebA_BlondHair}, we observe that none of the existing methods applied to CelebA can significantly reduce the bias $I(Z;A)$ in the extreme bias setting. Similarly, in~\cref{tab:Adult_mostEx}, we observe that none of the methods applied to Adult dataset can reduce $I(Z;A)$. These observations are explained by~\cref{th:extreme} which states that maintaining classification performance above random guess while achieving $I(Z;A)=0$ at $H(Y|A)=0$ is impossible.

In~\cref{tab:CelebA_BlondHair}, we also observe a trade-off between $I(Z;A)$ and accuracy, where reducing the bias (when $\Delta > 0$) results in a reduction in accuracy. This is explained by~\cref{th:general}, which states that when $H(Y|A)=0$, the amount of remained bias in the learnt feature $Z$ is an upper bound to the best performance, \ie $I(Z;Y)\leq I(Z;A)$, and therefore removing more bias can result in lower performance. The only exception to this trade-off seems to be DI~\cite{domain_independent_training}. We conjecture that this is due to its enhanced ability in achieving the best attainable classification performance (see~\cref{th:best_att_perf}).

\subsection{Analysis of the Strong Bias Region $H(Y|A)>0$}
\label{subsec:main_comparison}
In this section, we go beyond the extreme bias point, and more generally investigate the consequences of applying existing bias removal methods on the entire range of bias strength, \ie connecting the extreme bias training setting (TrainEx) we studied in~\cref{sec:comp_extreme} to the moderate bias in the original training setting (TrainOri) commonly studied in existing methods. We again study two aspects of each method, its classification performance (measured by accuracy on Unbiased and Bias-conflicting settings) and its ability to remove bias (measured by estimating $I(Z;A)$ using~\cite{MINE} on the training set). In~\cref{fig:EB_CelebA,fig:EB_Adult}, we observe a decline in the performance of all methods as the bias becomes stronger, in both CelebA and Adult datasets, similar to our observation in Colored MNIST in~\cref{fig:EB}. This observation is consistent with~\cref{th:general}, which states that the bias strength determines an upper bound on the best performance of bias removal methods, regardless of the dataset and method.

In~\cref{fig:CelebA_IZA,fig:Adult_IZA}, we use breaking points to approximately divide the strong bias region into three phases and explain the observed changes in the performance of methods from the perceptive of~\cref{th:general}. In phase 1, as $H(Y|A)$ increases from zero to the breaking point (bias strength decreases), we observe that the attribute bias $I(Z;A)$ is not minimized because of the trade-off between best attainable performance $I(Z;Y)$ and attribute bias removal when bias is very strong: the methods choose to increase accuracy towards the best attainable accuracy $I(Z;Y)$ rather than removing attribute bias (this choice is most likely due to the larger weight on the accuracy term in their objectives). Then, in phase 2, as $H(Y|A)$ increases through the breaking point (bias strength decreases further), the methods start to minimize attribute bias $I(Z;A)$ because the upper bound on best attainable performance $I(Z;Y)$ is now large enough to avoid the trade-off between accuracy and attribute bias removal.
Finally, in phase 3, as $H(Y|A)$ further departs from the breaking point, accuracy gradually approaches its best attainable performance, while attribute bias $I(Z;A)$ is minimized further below that of the baseline because the weaker bias strength now allows the model to distinguish $Y$ from $A$ so that minimizing attribute bias and maximizing accuracy do not compete.

\section{Conclusion and Future Work}
\label{sec:discussion}
In this work, we mathematically and empirically showed the sensitivity of state-of-the-art attribute bias removal methods to the bias strength.
This highlights a previously overlooked limitation of these methods. In particular, we empirically demonstrated that when a protected attribute is strongly predictive of a target, these methods become ineffective. To understand the cause and extent of these findings, we derived an information-theoretical upper bound on the performance of any attribute bias removal method, and verified it in experiments on synthetic, image, and census datasets. These findings not only caution against the use of existing attribute bias removal methods in datasets with potentially strong bias (\eg small datasets), but also motivate the design of future methods that can work even in strong bias situations, for example by utilizing external unlabelled datasets to relax the upper bound. Additionally, investigating the role of bias strength in removing attribute bias from generative models is another interesting direction for future research.

\section*{Acknowledgement}

This research is based upon work supported in part by the Office of the Director of National Intelligence (ODNI), Intelligence Advanced Research Projects Activity (IARPA), via [2022-21102100007]. The views and conclusions contained herein are those of the authors and should not be interpreted as necessarily representing the official policies, either expressed or implied, of ODNI, IARPA, or the U.S. Government. The U.S. Government is authorized to reproduce and distribute reprints for governmental purposes notwithstanding any copyright annotation therein.


\bibliography{reference}
\bibliographystyle{ieee_fullname}

\clearpage
\appendix
\section{Proofs}
We consider all random variables to be discrete, as these are represented by floating point numbers in practice.
 
\noindent
\textbf{Proposition 1.} \emph{Given random variables $Z, Y, A$, in case of the extreme attribute bias $H(Y|A)=0$, if the attribute is removed from the feature $I(Z;A)=0$, then $I(Z;Y)=0$, \ie no classifier can outperform random guess.}

\begin{proof}
 
Generally, we have,
\begin{equation}
\begin{split}
	p(z,y) &= \sum_{a} p(z,y,a) \\
	&= \sum_{a} p(y|z,a) p(z,a).
\end{split}
\end{equation}
As $I(Z;A)=0$ and by the property of mutual information that $I(Z;A)=0$ if and only if random variable $Z$ and random variable $A$ are mutually independent, we have,
 
\begin{equation}
\begin{split}
\label{2}
	p(z,y) &= \sum_{a} p(y|z,a) p(z) p(a). \\
\end{split}
\end{equation}
By the property of conditional entropy that $H(Y|A)=0$ if and only if $\exists~\text{a function}~g: \mathcal{A} \to \mathcal{Y}, Y=g(A)$, and the property that if $Y$ is determined by $A$ with a function $g$, $\forall~\text{random variable}~Z, p(y|a,z) = p(y|a)$, \cref{2} goes to,

\begin{equation}
\begin{split}
	p(z,y) &= \sum_{a} p(y|a) p(z) p(a) \\
	&= \sum_{a} p(y,a) p(z) \\
	&= p(y) p(z).
\end{split}
\end{equation}
Thus, $Z$ and $Y$ are mutually independent, and we have,
\begin{equation}
\begin{split}
	I(Z;Y) = 0.
\end{split}
\end{equation}
\end{proof}

\newpage
 
\noindent
\textbf{Theorem 1.}
\emph{Given random variables $Z, Y, A$, the following inequality holds without exception:}
\begin{align}
	0 \leq I(Z;Y) \leq I(Z;A) + H(Y|A)
\end{align}

\begin{proof}
For interaction information $I(Z;Y;A)$, we have,
\begin{equation}
\begin{split}
I(Z;Y;A) &= I(Z;Y) - I(Z;Y|A).
\end{split}
\end{equation}
By the symmetry property of mutual information,
\begin{equation}
\begin{split}
I(Z;Y;A) = I(Z;Y) - I(Y;Z|A).
\end{split}
\end{equation}
By the chain rule of conditional mutual information,
\begin{equation}
\begin{split}
I(Z;Y;A) &= I(Z;Y) - [I(Y;Z,A) - I(Y;A)] \\
&= I(Z;Y) - I(Y;Z,A) + I(Y;A).
\end{split}
\end{equation}
By the property of interaction information $I(Z;Y;A) \leq \min\{I(Z;Y), I(Y;A), I(Z;A)\}$, we have,
\begin{equation}
\begin{split}
I(Z;Y;A) &\leq I(Z;A) \\
I(Z;Y) - I(Y;Z,A) + I(Y;A) &\leq I(Z;A).
\label{4}
\end{split}
\end{equation}
According to the relation of mutual information to conditional and joint entropy, Left Hand Side (LHS) of~\cref{4} goes to,
\begin{equation}
\begin{split}
LHS = &I(Z;Y) - [H(Y)-H(Y|Z,A)] \\
&+ [H(Y) - H(Y|A)] \\
= &I(Z;Y) + H(Y|Z,A) - H(Y|A).
\end{split}
\end{equation}
Then, \cref{4} goes to,
\begin{equation}
\begin{split}
I(Z;Y) + H(Y|Z,A) - H(Y|A) &\leq I(Z;A), \\
I(Z;Y) + H(Y|Z,A) &\leq I(Z;A) + H(Y|A).
\end{split}
\end{equation}
As $H(Y|X,A) \geq 0$ and mutual information is non-negative, we have,
\begin{equation}
\begin{split}
0 \leq I(Z;Y) &\leq I(Z;A) + H(Y|A).
\end{split}
\end{equation}
\end{proof}

\clearpage

\section{Statistics of Attribute Bias in Various Real-World Datasets}

In this section, we present detailed statistics of attribute bias in various real-world datasets to illustrate that the presence of strong bias is a common phenomenon in real-world datasets. 
For instance, in the healthcare domain, a publicly available diabetes dataset, Pima Indians Diabetes Dataset~\cite{diabetes_dataset}, collected by the National Institute of Diabetes and Digestive and Kidney Diseases, contains a large portion of negative samples for young individuals (352) and a small portion of negative samples for old individuals (148), which leads age to be strongly predictive of diabetes diagnoses~\cite{diabetes_chapter}.\footnote{Following~\cite{BlindEye_IMDB_eb}, we choose age around 30 (33) as the boundary to transform the origin continuous age label to be discrete (\ie old versus young).}
Moreover, diagnosing Human Immunodeficiency Virus (HIV) from Magnetic Resonance Imaging (MRI) images, by statistics in~\cite{dataset_vs_task}, age of control group is 45{\scriptsize$\pm$17.0} while the age of HIV subjects is 51{\scriptsize$\pm$8.3}, making age a very strong attribute bias for this task.
Furthermore, in CelebA dataset, \textit{HeavyMakeup}, which is a widely-used attribute to study attribute bias in~\cite{LfF_CelebA_Bias_conflicting, CSAD, End} is strongly predictive of sex.  
This is evident from the fact that the positive rate of HeavyMakeup in males is only $0.28\%$ (234 out of 84434), which is close to zero, whereas the positive rate of HeavyMakeup in females is $66\%$ (78156 out of 118165). Therefore, the classification of HeavyMakeup is highly influenced by the spurious association between HeavyMakeup and sex. Similarly, other non-sex-related attributes in CelebA also exhibit a significant disparity in the positive rate between males and females. For example, \textit{BlondHair} (2\%, 24\%), \textit{WavyHair} (14\%, 45\%), \textit{HighCheekbones} (31\%, 56\%), \textit{BagsUnderEyes} (35\%, 10\%), \textit{BigNose} (42\%, 10\%), and \textit{PointyNose} (16\%, 36\%).
All of these statistics illustrate that strong bias is a common problem in the real world.

\section{Connection with ``Impossibility results for fair representations''}

Lechner \etal~\cite{impossibility_for_fair_repesentations} claim that the non-trivial fair representations in terms of Demographic Parity and Odds Equality are unattainable.
Demographic Parity (DP), which requires independence between model predictions $\hat{Y}$ and attributes $A$ ($I(\hat{Y};A)=0)$, aligns with the bias issues discussed in our work.
Thus, we mainly clarify the connection with DP.

Specifically, Claim 1 in~\cite{impossibility_for_fair_repesentations} states that any representation that accommodates a non-constant classifier cannot ensure DP-fairness for all potential tasks involving the same set of individuals.
This claim can be proved by finding a task for which no representation can guarantee DP-fairness.
For example, they consider the task to predict attribute $A$ itself, rendering the fair representation with respect to DP for this particular task cannot exist.
However, it is important to note that the claim argues that DP-fair representation cannot exist for all potential tasks but it does not negate the feasibility of the specific task in general attribute bias scenarios.
Furthermore, the task they provide, where the target $Y$ is the attribute $A$ itself, aligns with the extreme bias case where $H(Y|A)=0$ in our paper.
In this scenario, as elaborated in~\cref{sec:theory}, we also claim that if $H(Y|A)=0$ and $I(Z;A)=0$, we have $I(Z;Y)=0$, \ie no classifier can outperform random guess.

\section{Breaking Point of Attribute Bias Removal Methods on Colored MNIST}
\label{sec:breaking_point}

In this section, we present the details regarding breaking point in~\cref{fig:EB} of the main paper.

As mentioned in~\cref{sec:introduction} of the main paper, digit classification on Colored MNIST, assuming color as a protected attribute to be removed, is a popular controlled experiment used by several attribute bias removal methods~\cite{learn_not_to_learn_Colored_MNIST, Back_MI, CSAD} to measure their effectiveness. More concretely, this is a ten-class classification problem given an image as input, where in the training set each digit is assigned a unique RGB color with a fixed variance across all digits, and then to measure how much the trained model is relying on color to predict digit, its accuracy is reported on a testing set where colors are assigned to each sample at random. The higher the accuracy, the less the model has learnt to rely on the color -- the protected attribute in training -- for predicting digits. The experiment is repeated for different values of the color variance, and the results are plotted against color variance to show the sensitivity of a method to different levels of bias strength. While existing methods are effective in this experiment -- with a minor sensitivity to bias strength -- the performance is always only reported in the color variance range of $[0.02, 0.05]$, without any explicit justification on the choice of this range, thus leaving the question of whether these methods are effective outside of this range open. 

To empirically address this question, we extend the Colored MNIST experiment in two ways. 
First, we extend the horizontal axis to zero to include the color variance region $[0,0.02]$ where color becomes very predictive of the digit, hence the attribute bias strength is increased to its maximum. We denote this range the \textit{strong bias region}, with extreme bias happening at zero. \cref{fig:EB} in the main paper shows that unlike in the moderate bias region, the performance of all existing methods becomes very sensitive to the bias strength in the strong bias region. Moreover, we see that in extreme bias, all methods perform close to chance. 
Second, we repeat the experiment at each color variance 15 times and compute error bars for the whole range. This enables us to not only visually judge the significance of differences in performance, but perhaps more interestingly, be able to compute a \textit{breaking point} for each method. Formally, we define the breaking point of a method as the largest color variance (smallest bias strength) where its performance is no longer significantly better than a naive baseline. The naive baseline only uses the typical cross-entropy loss, without any implicit or explicit mutual information minimization term, and otherwise has the exact same classification network structure as all the other methods. To compute the breaking point, we conduct hypothesis tests based on a two-sample one-way Kolmogorov–Smirnov test at different color variances and set the null hypothesis to that model performance is better than baseline performance: the largest color variance at which the null hypothesis is rejected with a $p$-value that is less than or equal to the significance level of 0.05 is considered as the breaking point. 
We provide the $p$-values of each method over different color variances in~\cref{tab:Pvalue}.
In~\cref{fig:EB} of the main paper, the breaking point of each method is illustrated with a $\blacktriangle$ on the x-axis. Interestingly, we observe that different methods have different breaking points, showing that methods that might appear to perform very closely in the moderate bias region (\eg compare LNL and EnD), can perform very differently in the strong bias region.

These empirical findings reveal that the existing methods are only effective under the previously unstated assumption that the attribute bias is not too strong, an important consideration for the use of these methods in practice. But equally important, we observe a clear connection between the effectiveness of existing methods and the attribute bias strength. 
Then, in~\cref{sec:theory} of the main paper, we formulate this connection.
 
\begin{table*}[htbp]
\caption{$p$-value of hypothesis tests over different color variances on Colored MNIST. The significant level of hypothesis tests is 0.05. Color variance is consecutive in its three rows. \textbf{Bold} for the $p$-value on the breaking point of each method. While existing methods are effective over baseline in moderate bias region where color variance exceeds 0.02, they perform close to or worse than baseline in strong bias region and own different breaking points.}

\label{tab:Pvalue}
\centering

\begin{tabular}{lrrrrrrrrr}
\toprule
Color variance            & 0.000                & 0.001                & 0.002                & 0.003                & 0.004                & 0.005                & 0.006                & 0.007                & 0.008                \\
\midrule
LNL~\cite{learn_not_to_learn_Colored_MNIST}          & 0.025                & 0.044                & 0.000                & 0.000                & 0.048                & 0.040                & 0.032                & 0.025                & 0.040                \\
BackMI~\cite{Back_MI}        & 0.034                & 0.020                & 0.043                & 0.010                & 0.023                & 0.037                & 0.042                & \textbf{0.036}       & 0.540                \\
EnD~\cite{End}         & 0.043                & 0.000                & 0.042                & 0.016                & 0.017                & 0.048                & 0.025                & 0.016                & 0.018                \\
CSAD~\cite{CSAD}          & 0.038                & 0.048                & \textbf{0.040}       & 0.723                & 1.000                & 1.000                & 1.000                & 1.000                & 1.000                \\
\midrule
\midrule
Color variance           & 0.009                & 0.010                & 0.011                & 0.012                & 0.013                & 0.014                & 0.015                & 0.016                & 0.017                \\
\midrule
LNL~\cite{learn_not_to_learn_Colored_MNIST}          & \textbf{0.045}       & 0.436                & 0.990                & 1.000                & 0.970                & 0.970                & 1.000                & 1.000                & 1.000                \\
BackMI~\cite{Back_MI}       & 0.960                & 1.000                & 1.000                & 0.990                & 0.960                & 0.990                & 1.000                & 0.980                & 1.000                \\
EnD~\cite{End}          & 0.044                & 0.034                & 0.018                & \textbf{0.040}       & 0.999                & 0.999                & 0.996                & 0.998                & 0.999                \\
CSAD~\cite{CSAD}          & 1.000                & 1.000                & 1.000                & 1.000                & 1.000                & 1.000                & 1.000                & 1.000                & 1.000                \\
\midrule
\midrule
Color variance           & 0.018                & 0.019                & 0.020                & 0.025                & 0.030                & 0.035                & 0.040                & 0.045                & 0.050                \\
\midrule
LNL~\cite{learn_not_to_learn_Colored_MNIST}          & 0.999                & 1.000                & 1.000                & 0.999                & 1.000                & 1.000                & 1.000                & 1.000                & 0.998                \\
BackMI~\cite{Back_MI}       & 0.996                & 1.000                & 1.000                & 1.000                & 1.000                & 1.000                & 1.000                & 1.000                & 1.000                \\
EnD~\cite{End}          & 0.994                & 1.000                & 1.000                & 1.000                & 1.000                & 1.000                & 1.000                & 1.000                & 0.999                \\
CSAD~\cite{CSAD}         & 1.000                & 1.000                & 1.000                & 1.000                & 1.000                & 1.000                & 1.000                & 1.000                & 1.000                \\
\bottomrule
\end{tabular}

\end{table*}



\clearpage

\end{document}